\pdfoutput=1
\documentclass[11pt,a4paper,dvipsnames]{article}
\usepackage[hyperref]{acl}
\usepackage{times}
\usepackage{latexsym}
\usepackage{graphicx}
\usepackage[]{xcolor}
\usepackage{cleveref}
\usepackage{array,booktabs,ragged2e}
\newcolumntype{P}[1]{>{\centering\arraybackslash}p{#1}}
\newcolumntype{M}[1]{>{\centering\arraybackslash}m{#1}}
\newcolumntype{R}[1]{>{\RaggedLeft\arraybackslash}p{#1}}

\usepackage[T1]{fontenc}

\newcommand{\Ni}{({\em i})~}
\newcommand{\Nii}{({\em ii})~}
\newcommand{\Niii}{({\em iii})~}

\graphicspath{{img/}}

\usepackage{microtype}
\usepackage{multirow,tabularx}

\title{OCR Improves Machine Translation for Low-Resource Languages} %

\author{Oana Ignat$^1$\hspace{3mm} Jean Maillard$^2$\hspace{3mm} 
{\bf Vishrav Chaudhary}$^3$\thanks{\ \ Work done while at Meta AI.}\footnotemark[1] \hspace{2mm} {\bf Francisco Guzmán}$^2$\\ 
University of Michigan$^1$ \hspace{1em} Meta AI$^2$ \hspace{1em}
Microsoft Turing$^3$\\
  \texttt{oignat@umich.edu}, \texttt{jean@maillard.it}, \texttt{vchaudhary@microsoft.com} \\
  \texttt{fguzman@fb.com} 
}

\begin{document}
\maketitle
\begin{abstract}
We aim to investigate the performance of current OCR systems on low resource languages and low resource scripts.
We introduce and make publicly available a novel benchmark, \textsc{OCR4MT}, consisting of real and synthetic data, enriched with noise, for 60 low-resource languages in low resource scripts. We evaluate state-of-the-art OCR systems on our benchmark and analyse most common errors. We show that OCR monolingual data is a valuable resource that can increase performance of Machine Translation models, when used in backtranslation. We then perform an ablation study to investigate how OCR errors impact Machine Translation performance and determine what is the minimum level of OCR quality needed for the monolingual data to be useful for Machine Translation.
\end{abstract}

\section{Introduction}

Despite many recent successes, Machine Translation still lacks support or fails to achieve good performance for most low-resource languages, which represent a very large fraction of the languages spoken by the world's population \cite{fan2020BeyondEM, Wenzek2020CCNetEH, Goyal2021TheFE}.

The poor performance in these settings can largely be attributed to the lack of training data. Many techniques for improving Machine Translation, such as backtranslation \cite{Sennrich2016ImprovingNM,edunov-etal-2018-understanding,zhang-etal-2020-improving} and approaches which make use of pre-trained language models \cite{gao-etal-2019-soft,chen-etal-2021-zero,liu-etal-2021-counterfactual}, rely heavily on high quality monolingual data, which is not readily available for low-resource languages. Fortunately, many books and other resources in these languages have been digitized and made available online. However, this textual data is ``locked'' away in formats such as PDFs and images, which are not readily accessible. As a result, there are large unexplored collections of data in many languages which could be used as a source for monolingual data. For example, one Nepali books corpus\footnote{\url{https://pustakalaya.org/en/}}, contains around 342M tokens, which would potentially make it one of the largest sources of monolingual data for this language. %

A solution to this problem is to %
rely on modern Optical Character Recognition (OCR) tools to extract the text. Unfortunately however, most of the OCR models have only been evaluated on a handful of languages, and public benchmarks for low-resource scripts and languages are lacking \cite{tesseract,wick_calamari_2020}. As a result, a comprehensive evaluation of OCR tools, particularly for low-resource languages and scripts, is still an open problem. Moreover, there is little-to-no understanding of the downstream effect that recognition errors will have on the data augmentation techniques that make use of high-quality monolingual data, such as the methods that low-resource language translation typically relies upon.

In this paper, we pose the question of what is the minimum level of OCR quality needed for OCR-extracted monolingual text to be useful for Machine Translation, particularly in low-resource scenarios. To this end, in this work: \Ni we create and release an OCR benchmark, \textsc{OCR4MT}, first of its kind, based on real and synthetic data, enriched with noise, for 60 low-resource languages in low resource scripts; \Nii we evaluate commercial and research state-of-the-art OCR models on our benchmark, analyse their performance and extract their common errors for many languages; and \Niii we investigate how the most frequent OCR errors impact Machine Translation performance and determine what is the minimum level of OCR quality needed for monolingual data to be useful for Machine Translation.

From our results, %
we observe that the best available OCR systems work well on Latin scripts and perform significantly worse on non-Latin and non-European scripts (e.g., Perso-Arabic, Khmer). 

Our findings also show that monolingual data from OCR is a valuable source of data for improving Machine Translation for low resource languages, paving the way for future research on data augmentation for Machine Translation based on monolingual data extracted from OCR-ed documents.

\section{Related Work}

Despite extensive progress, Machine Translation for low resource languages is still an unsolved problem. 
This is mainly due to two different aspects: model architecture and lack of training data. In our work we focus on addressing the latter aspect.

One effective method to increase training data is to augment the parallel training corpus with backtranslations of target language sentences \cite{Sennrich2016ImprovingNM, edunov-etal-2018-understanding}.

There are large collections of unexplored scanned documents (i.e., PDFs) and images in low resource languages, that can be used as monolingual data for backtranslation, such as online repositories of books\footnote{\url{https://pustakalaya.org/en/}} or online archives\footnote{\url{https://archive.org/}}. Works like \citet{Rijhwani2020OCRPF} or \citet{Bustamante2020NoDT} also acknowledge that textual data for most low-resource languages often exists in formats that are not machine-readable, such as paper books and scanned images. They address the task of extracting text from these resources and create benchmark datasets of transcriptions for several endangered languages: Ainu, Griko, Yakkha  \cite{Rijhwani2020OCRPF} and Shipibo-konibo, Ashaninka, Yanesha, Yine \cite{Bustamante2020NoDT}. A summary of current benchmarks and data resources for low-resource languages in \Cref{tab:benchmarks_related}.
Observe that the related benchmarks contain few languages and few data compared to ours. 

Our research can be applied on large data resources of endangered and low resource languages, such as  AILLA\footnote{\url{https://ailla.utexas.org/}} or
ELAR\footnote{\url{https://elar.soas.ac.uk/}}.
\citet{Rijhwani2020OCRPF} find that endangered language linguistic archives contain thousands of scanned documents — the Archive of the Indigenous Languages of Latin America (AILLA) contains around 10,000 such documents and the Endangered Languages Archive (ELAR) has around 7,000. \citealt{Rijhwani2020OCRPF} find that endangered language documents often contain a translation into another (usually high-resource) language. Multilingual documents represent the majority in
the archives they examined: AILLA contains 4,383 scanned documents with bilingual text and 1,246 scanned documents with trilingual text, while ELAR contains around 5,000 multilingual documents.

This monolingual data can be collected using Optical Character Recognition (OCR) tools. However, we don’t know what is the quality of OCR tools, particularly for low-resource languages and low resource scripts. We aim to address this problem, by building a benchmark of 60 low resource languages with the goal of testing OCR systems and analyse how their errors impact backtranslation performance.

\citealt{Rijhwani2020OCRPF} also show how general-purpose OCR tools such as \cite{Fujii2017SequencetoLabelSI, Ingle2019ASH} are not robust to the data-scarce setting of endangered languages. They address this problem, by developing an OCR post-correction method tailored to ease the training in this data-scarce setting.

The work most similar to ours is the recent research by \citealt{Gupte2021LightsCA}. They also built a pipeline to generate analog synthetic documents on which they run a commercial OCR model and analyse the OCR errors. Unlike our work, however, their focus is on improving Named Entity Recognition (NER) accuracy and on only 4 different languages: (English, German) from CoNLL 2003 \cite{Sang2003IntroductionTT} and (English, Chinese and Arabic) from CoNLL 2012 \cite{Pradhan2012CoNLL2012ST}.

Our work's novelty consists in providing the first large-scale benchmark of 60 low resource languages and low resource scripts, with the purpose of evaluating OCR performance on each language and it's downstream impact on Machine Translation.

\begin{table}
\centering
\scriptsize
\setlength{\tabcolsep}{0.3em} %
{\renewcommand{\arraystretch}{1.3}%
\begin{tabular}{ M{0.22\textwidth} r r } 
    \toprule
     & \textbf{\#languages} & \textbf{ \#lines} \\
    \midrule
    \citet{Rijhwani2020OCRPF}  &  3 & 1,782 \\ 
    \citet{Bustamante2020NoDT} &  4 & 60,000  \\ 
    \citet{Gupte2021LightsCA} & 4 & not specified \\
    \textsc{OCR4MT} & 60 & 186,060 \\
    \bottomrule
\end{tabular}
}
\caption{Summary of some current benchmarks for low resource and endangered languages.}

\label{tab:benchmarks_related}
\end{table}

\section{\textsc{OCR4MT} Benchmark}
To build a benchmark useful for multiple low-resource languages and low resource scripts, we proposed the use of texts that are freely-available in multiple languages. To this end, 
we chose the Universal Declaration of Human Rights (UDHR) database\footnote{\url{https://www.unicode.org/udhr/translations.html}} which represents a legal domain, and the Flores 101 dataset \cite{Goyal2021TheFE} which is based on Wikipedia. Moreover, we chose these datasets because they provide data in many languages, and have plain text we can evaluate OCR models on. Our benchmark contains \textit{real} and \textit{artificially-created} PDFs\footnote{We call \textit{real}, the PDFs that had this format originally and we call \textit{artificially-created}, the PDFs that were originally text documents and were converted into PDF format. We \textit{artificially created} PDFs in order to
increase our benchmark data size, as by applying augmentation techniques (i.e., adding noise) they can resemble the \textit{real} PDFs.}. 

UDHR is composed of articles on fundamental human rights to be universally protected and it has been translated into over 500 languages. For each language, UDHR contains documents in different formats: plain text, PDF, XML and HTML. There are currently 460 translations fully converted to Unicode and available as text. Each document is composed of 30 short articles, on average 3 sentences each. We used the plain text and corresponding PDF files as validation data for the OCR systems.

The Flores 101 dataset consists of text data: 3,001 sentences extracted from English Wikipedia, for 101 languages, covering a variety of different topics and domains. We artificially created PDFs from the text documents by saving/exporting the text documents as PDF.

\paragraph{Language Selection.} We select 60 languages which are both in Flores 101 and the UDHR datasets. We prioritize low resource languages, with low resource scripts. The scripts, together with the corresponding languages present in our benchmark can be seen in \Cref{tab:scripts_langs}.

\begin{table}[hbt!]
    \centering
    \scriptsize
    {\renewcommand{\arraystretch}{1.3}%
    \begin{tabular}{ p{2.5cm}p{4cm} } 
    \toprule
     \textbf{Scripts} & \textbf{Languages} \\
    \midrule
    \multicolumn{2}{c}{\textbf{Latin}} \\
    \midrule
     Latin & Asturian, Cebuano, Fula, Ganda, Icelandic, Lingala, Maori, Nyanja, Oromo, Polish, Portuguese (Portugal), Romanian, Shona, Slovak, Slovenian, Somali, Swahili, Swedish, Turkish, Umbundu, Uzbek, Vietnamese, Wolof, Zulu \\
    \midrule
    \multicolumn{2}{c}{\textbf{Cyrillic}} \\
    \midrule
    Cyrillic & Belarusian, Bulgarian, Kazakh, Kyrgyz, Macedonian, Mongolian, Russian, Serbian, Tajik, Ukrainian \\
    \midrule
    \multicolumn{2}{c}{\textbf{Perso-Arabic}} \\
    \midrule
    Arabic & Arabic, Sorani Kurdish \\
    Perso-Arabic & Pashto, Urdu \\
    \midrule
    \multicolumn{2}{c}{\textbf{North Indic}} \\
    \midrule
    Bengali & Bengali \\
    Devanagari & Hindi, Marathi, Nepali\\
    Gujarati & Gujarati \\
    Gurmukhi & Punjabi \\
    \midrule
    \multicolumn{2}{c}{\textbf{South Indic}} \\
    \midrule
    Malayalam & Malayalam \\
    Tamil & Tamil \\
    Telugu-Kannada & Kannada, Telugu \\
    \midrule
    \multicolumn{2}{c}{\textbf{Southeast Asian (SEA)}} \\
    \midrule
    Khmer & Khmer \\
    Lao & Lao \\
    Myanmar & Burmese \\
    Thai & Thai \\
    \midrule
    \multicolumn{2}{c}{\textbf{China-Japan-Korea (CJK)}} \\
    \midrule
    Han& Japanese \\
    Hangul & Korean \\
    Hant & Chinese Simpl \\
    \midrule
    \multicolumn{2}{c}{\textbf{Others}} \\
    \midrule
    Armenian & Armenian \\
    Ge’ez & Amharic \\
    Georgian & Georgian \\
    Greek & Greek \\
    Hebrew & Hebrew \\
    \bottomrule
    \end{tabular}
    }
\caption{Scripts and their corresponding languages in our benchmark. The languages are grouped into 8 groups, according to their location and script.}
    \label{tab:scripts_langs}
\end{table}

\paragraph{Annotation Process.} 
The UDHR data is composed of one document image per language (PDF), and each document contains a preface and around 30 articles. In addition, each document has an accompanying text version. To build the benchmark, we first manually annotate the bounding boxes for each of the 30 PDF documents. Using the bounding boxes, we split each document image into individual articles of about 40 words in average. This allows to accurately compare the ground truth text version with the OCR output for each article. %

Each article was labeled by a single annotator. We had a total of 10 annotators in total. In the tutorial we showed how to crop a bounding box around each article and how to name the images with their corresponding language code and number. 

\paragraph{Data validation.} We then validate the quality of annotations, both automatically and manually. We automatically validate each article by measuring the CER per article. If the CER between the PDF labeled version and the text version is greater than two standard deviations away from the mean, the article is marked as anomalous \cite{Cousineau2010OutliersDA}. We manually check and re-annotate all the anomalous articles until no anomalies were detected. 

During the manual anomaly check process, we found cases when for some languages, i.e., Malayalam and Pashto, some articles were missing in the original PDF document. In such cases, we removed those articles from the benchmark. We also found and removed all articles for which the PDF and text versions had different contents (i.e., they were paraphrases of each other). In total, we removed 141 articles, which is $\sim$7.8\% of the total number of initial articles.  Finally, we obtain 1,659 pairs of PDF and corresponding text versions of articles.

\begin{figure}[hbt]
    \centering
    \includegraphics[width=1\linewidth]{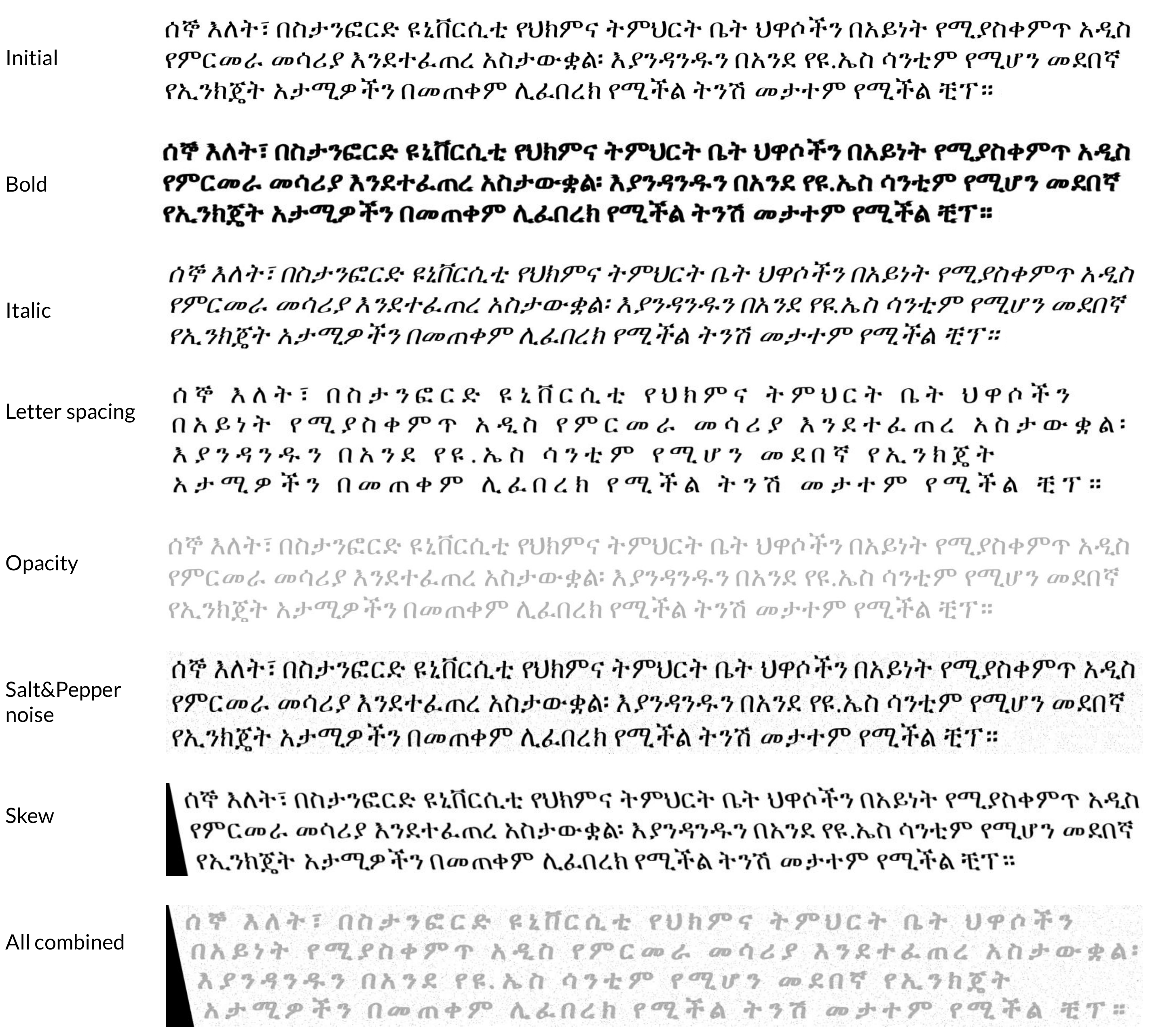}
    \caption{Data augmentation sample on Amharic artificial PDF from Flores 101: adding bold, italic, increasing letter spacing, decreasing opacity, adding salt and pepper, skewing and all combined.}
    \label{fig:augm}
\end{figure}

\paragraph{Data Augmentation.} To make the  artificial data closer to real life PDFs, we apply different augmentation techniques: changing font, color, size, letter spacing, opacity, italic, bold and image: skewing, adding salt \& pepper noise. We choose common fonts for the data scripts: Times New Roman (for Arabic, Latin), Arial (for Arabic, Cyrillic), Verdana (for Cyrillic), Noto Sans Devanagari (for Devanagari), Calibri (for Pashto), Jameel Noori Nastaleeq (for Urdu), Browalia New (for Thai), Korean (for Korean), PMingLiu (for Traditional Chinese). The letter spacing, opacity, skewing and noise levels can be adjusted. A sample augmented document from Flores 101 is shown in \Cref{fig:augm}.

\section{OCR Evaluation}
To estimate the impact of recognition errors in downstream tasks, namely Machine Translation, we perform a \emph{black-box} evaluation of two SOTA OCR systems, one commercial and one research. These represent reasonable choices for an non-OCR expert, such as MT practitioners. Below, we describe our experimental setup in detail.%

\subsection{OCR SOTA systems}
Following \citet{rijhwani-etal-2020-ocr}, for the commercial use case, we evaluate the Google Vision API OCR system \cite{Fujii2017SequencetoLabelSI, Ingle2019ASH} as provided by the Google Vision AI toolkit\footnote{\url{https://cloud.google.com/vision}}. For the research system, we use the Tesseract OCR engine \cite{Smith2007AnOO}.

Google Vision OCR system is highly performant and covers 60 major languages in 29 scripts. It also provides script-specific OCR models in addition to language specific ones. Per-script models are more robust to unknown languages because they are trained on data from multiple languages and can act as a general character recognizer without relying on a single language’s model \cite{Rijhwani2020OCRPF}.

Tesseract is one of the most accurate open-source OCR engines \cite{Smith2007AnOO}. In our experiments, we run Tesseract version 4, which is based on an LSTM architecture \cite{hochreiter1997long}.
Tesseract can recognize more than 100 languages and it can be trained to recognize other languages.

\subsection{Metrics}
The metrics we use for measuring OCR performance is character error rate (CER)  \cite{BergKirkpatrick2013UnsupervisedTO, Schulz2017MultimodularDO}. The metrics are based on the Levenshtein or edit distance, which is the minimum number of single-character edits (insertions, deletions or substitutions) required to change one word into the other. CER is the edit distance between the OCR-ed data and the gold standard/initial data, divided by the total number of characters in the initial data. CER is not always between 0 and 100, in particular when there is a high number of insertions. This value is often associated to the percentage of characters that were incorrectly predicted. 

Word error rate (WER), CER's word-based counterpart, is also used in related work  \cite{Rijhwani2020OCRPF, Rigaud2019ICDAR2C, Chiron2017ICDAR2017CO}. In this work, we choose to report only CER, as word boundaries are not comparable across languages.

There is no single benchmark for defining a \textit{good} CER value, as it is highly dependent on the use case. Different scenarios and complexity (e.g., printed vs. handwritten text, type of content, etc.) can result in varying OCR performances.
In \citet{Holley2009HowGC}, a review of OCR accuracy in large-scale Australian newspaper digitization programs came up with these benchmarks, for printed text:
\begin{itemize}
  \setlength\itemsep{-0.1em}
\item \textbf{Good} OCR accuracy: CER 1‐2\% (i.e., 98–99\% accurate)
\item \textbf{Average} OCR accuracy: CER 2-10\%
\item \textbf{Poor} OCR accuracy: CER $>$ 10\% (i.e., below 90\% accurate)
\end{itemize}

\begin{table*}[hbt!]
  \centering
  {\renewcommand{\arraystretch}{1.2}%
    \scriptsize
  \begin{tabular}{l R{2cm} R{3cm} R{2cm}  R{3cm}  }
   \toprule
    \textbf{OCR accuracy} & \multicolumn{2}{c}{\textbf{Flores 101}} & \multicolumn{2}{c}{\textbf{UDHR}} \\
    \cmidrule(r){4-5} \cmidrule(r){2-3}
      & Tesseract & \citealt{Fujii2017SequencetoLabelSI} & Tesseract & \citealt{Fujii2017SequencetoLabelSI} \\
    \midrule
    Good (CER $<$ 2\%) &60\% & 80\% & 35\% & 50\%\\
    Average (CER 2-10\%) & 28.3\% & 15\% & 31.7\% & 23.3\%\\
    Poor (CER $>$ 10\%) & 11.6\% & 5\% & 33.3\% & 26.7\% \\
 \bottomrule
  \end{tabular}
  \caption{Evaluation of SOTA models on our benchmark: percentage of languages with a good, average and poor OCR accuracy, on artificial PDFs (Flores 101) and real PDFs (UDHR).}
   \label{tab:eval_SOTA_summary}
  }
 
\end{table*}
\subsection{General Results}
We evaluate each model %
on the 60 languages from our benchmark, on both artificially created PDFs (Flores 101) and real PDFs (UDHR). 

From the results in \Cref{tab:eval_SOTA_summary}, we can see that the commercial system from \citet{Fujii2017SequencetoLabelSI} performs overall better than Tesseract across languages and data types:
20\% more languages have good performance on artificial data and 15\% more languages have good performance on real data.
In \Cref{tab:eval_SOTA} we also provide the results for each language, OCR system and data. 

\begin{figure}[hbt!]
    \centering
    \scalebox{.9}{
    \includegraphics[width=1\linewidth]{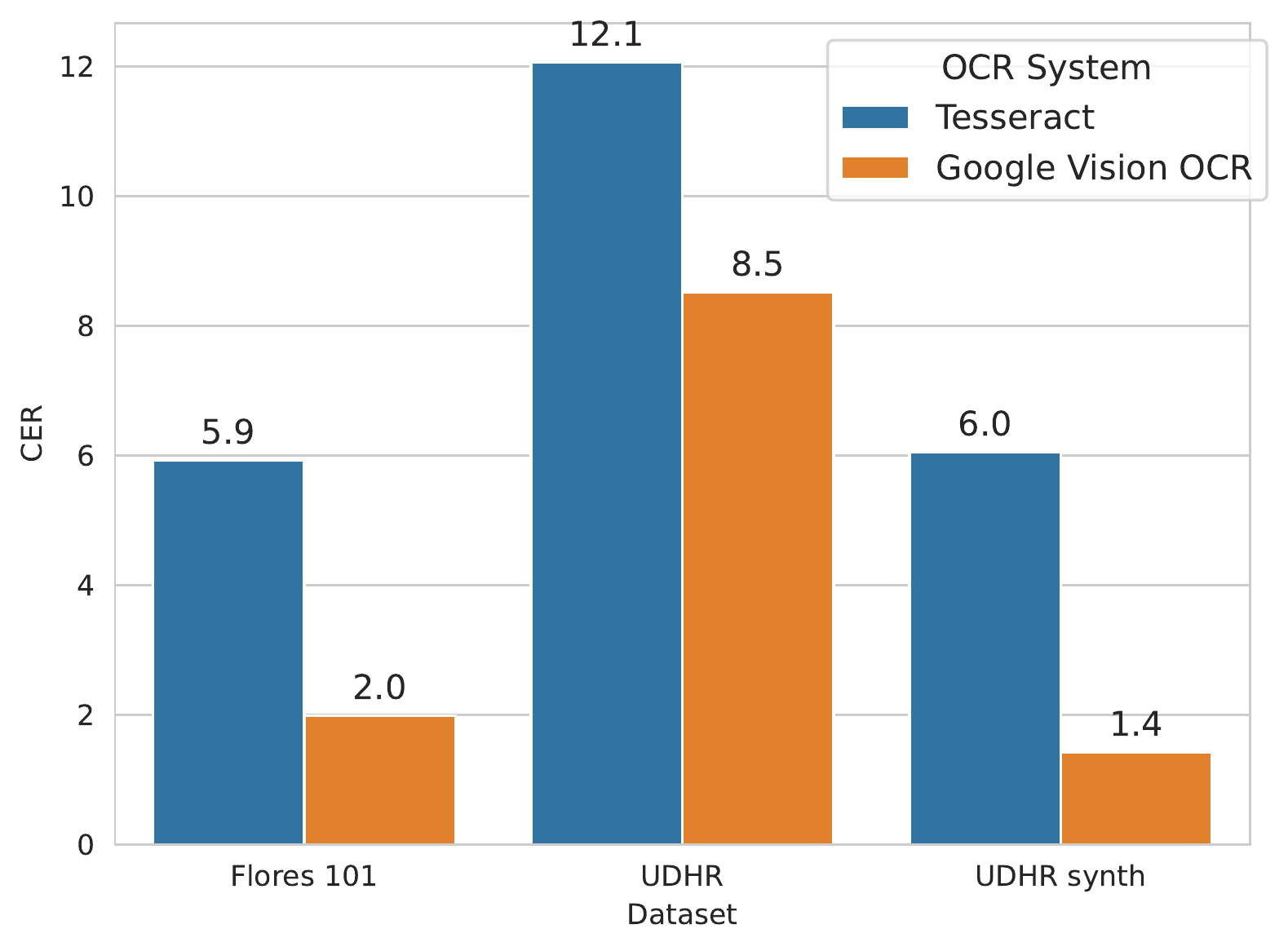}}
    \caption{Average CER (the lower, the better) of the SOTA OCR systems: Tesseract and \citealt{Fujii2017SequencetoLabelSI}, across datasets, over 60 languages. UDHR synth contains artificially created PDFs from UDHR.}
    \label{fig:eval_datasets}
\end{figure}

As expected, we also observe that the OCR performance is higher on artificially created PDFs (average CER 5.9 and 2.0) compared to real PDFs (average CER 12.1 and 8.5). We want to verify this is not due to the content, but to the format of the data. Therefore, we create artificial PDFs from the real ones in UDHR data, and run the OCR models on each of the 3 datasets. The results can be seen in \Cref{fig:eval_datasets}.

\subsection{Group analysis}
\begin{figure}[hbt!]
    \centering
    \includegraphics[width=1\linewidth]{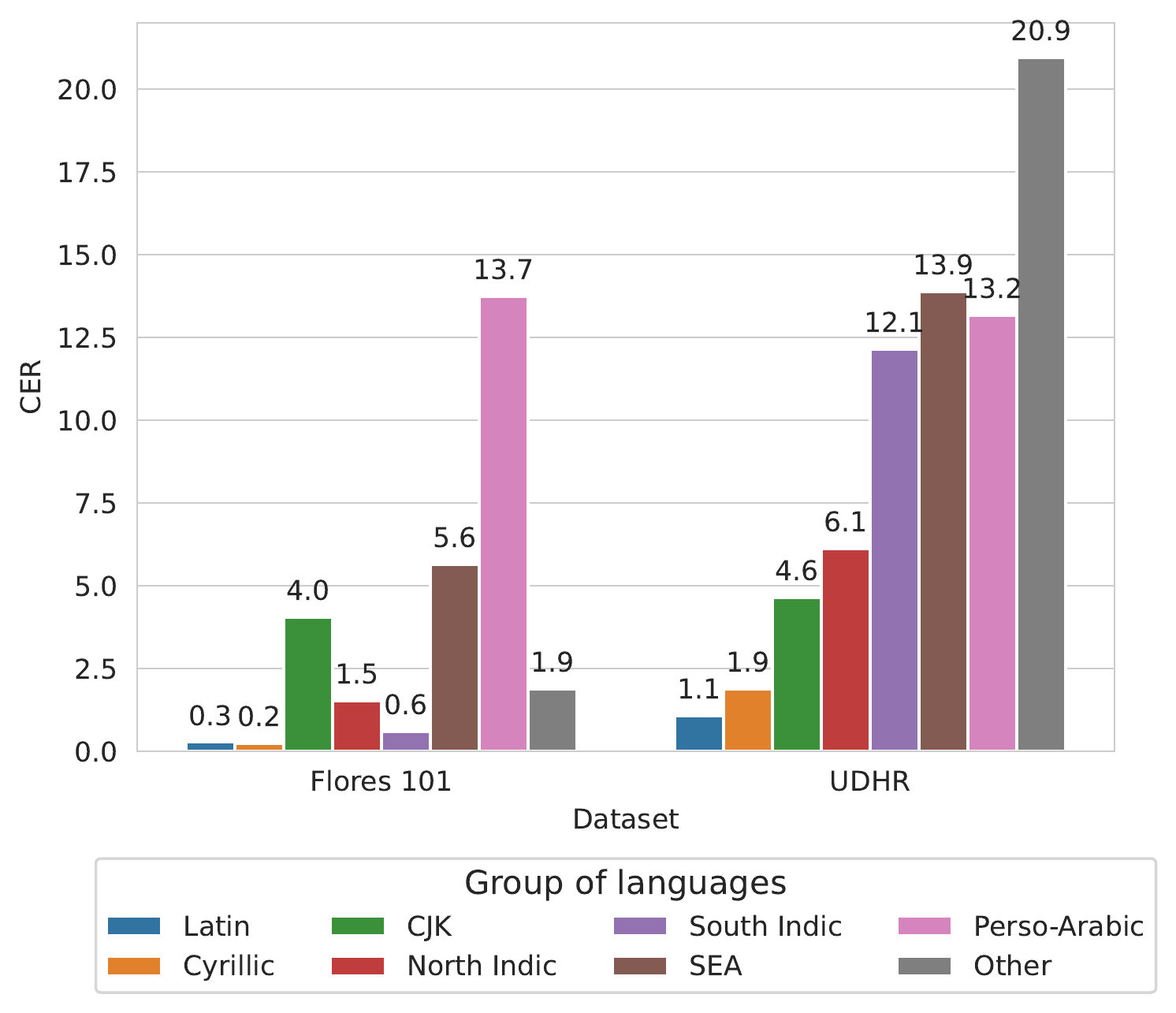}
    \caption{Average CER (the lower, the better) of best performing OCR model (\citet{Fujii2017SequencetoLabelSI}), across groups of languages in UDHR and Flores 101 datasets.}
    \label{fig:eval_groups}
\end{figure}

We also observe that the performance of the OCR systems vary based on script and location.
Therefore, we group the 60 languages into 8 groups, as in \Cref{tab:scripts_langs}, according to their script and location: Latin, Cyrillic, Perso-Arabic, North Indic, South Indic, Southeast Asian (SEA), China-Japan-Korea (CJK) and Other/Unique (Armenian, Amharic, Georgian, Greek, Hebrew).
We run the overall best OCR system \cite{Fujii2017SequencetoLabelSI} on these 8 groups of languages and compare the performance between language groups and also between the different data types: real PDFs (UDHR) and artificial PDFs (Flores 101).
The results can be seen in \Cref{fig:eval_groups}. Our observations and takeaways from this evaluation are the following:

\begin{itemize}
  \setlength\itemsep{-0.1em}
    \item  \textbf{Artificially created data is easier to recognize}. As expected, the OCR SOTA model performs overall better on artificially created PDFs (Flores 101) than on real PDFs (UDHR). This holds for each group of languages, with the exception of the Perso-Arabic group where the OCR accuracy is slightly poorer (13.7 CER on Flores 101 and 13.2 CER on UDHR). 
    \item \textbf{Latin and Cyrillic achieve the best performance}. The OCR SOTA model accuracy is the highest for European scripts such as Latin and Cyrillic. The OCR accuracy on Latin and Cyrillic is good ($<$ 2\% CER) on both Flores 101 and UDHR data. Therefore, we conclude that efforts for improving OCR models should focus on 
    groups of languages other than Latin and Cyrillic.
    \item \textbf{Perso-Arabic performs badly}. Given that the Perso-Arabic group has a poor performance on both Flores 101 and UDHR data ($>$ 10\% CER),  we conclude that the Perso-Arabic group needs considerable attention when improving OCR models.
    \item  \textbf{Performance varies per languages/type of data}. The North Indic, South Indic, SEA and Other/Unique (Armenian, Amharic, Georgian, Greek, Hebrew) groups have a good or average OCR accuracy on artificially created data (Flores 101) and a poor OCR accuracy on real data (UDHR). This shows that OCR models need more real training data from the North Indic, South Indic, SEA and Other/Unique (Armenian, Amharic, Georgian, Greek, Hebrew) groups. A notable exception is the performance for the CJK group, which has a similar performance on both datasets.

\end{itemize}

\vspace{-3mm}
\section{OCR impact in Machine Translation}

 Monolingual data is a valuable resource for Machine Translation, particularly for data augmentation techniques such as backtranslation. While there is plenty of monolingual data available for a few languages, there is a lack of data for very low resource languages. Fortunately, we have observed that there exist collections of monolingual data for low resource languages available as PDFs and images. 

However, we still do not know whether the quality of the OCR-ed data is good enough to be used for training and improve the performance of a Machine Translation (MT) model. In this section, we explore the performance of an MT model after being trained on backtranslated OCR-ed (OCR+BT) data. In particular, we explore the setup in which a pre-trained multilingual model is fine-tuned on backtranslated data obtained from OCR-ed monolingual data. We use this setup to understand the cases in which OCR data improves or hurts performance.
\subsection{The Nepali case}
One of the languages with a promising number of documents is Nepali, which contains around 342M tokens from the corpus of Nepali books\footnote{\url{https://pustakalaya.org/en/}}, which potentially makes it the largest sources of monolingual data for this language. To understand how valuable is the data and the validity of our evaluation setup, we explore adding OCR+BT data in small increments.

\paragraph{Setup.} We collect the OCR-ed Nepali data using the open-source model Tesseract \cite{Smith2007AnOO}. 
We then perform backtranslation, where we translate the OCR-ed Nepali data into English synthetic data using a SOTA MT model and use the data to fine-tune the model.
As SOTA MT model, we use the pre-trained model M2M-124 with 615M parameters from \citealt{Goyal2021TheFE} which was extended to 124 languages from the M2M-100 multilingual model \cite{fan2020BeyondEM}. 

\begin{figure}[hbt!]
    \centering
     \scalebox{.9}{
    \includegraphics[width=1\linewidth]{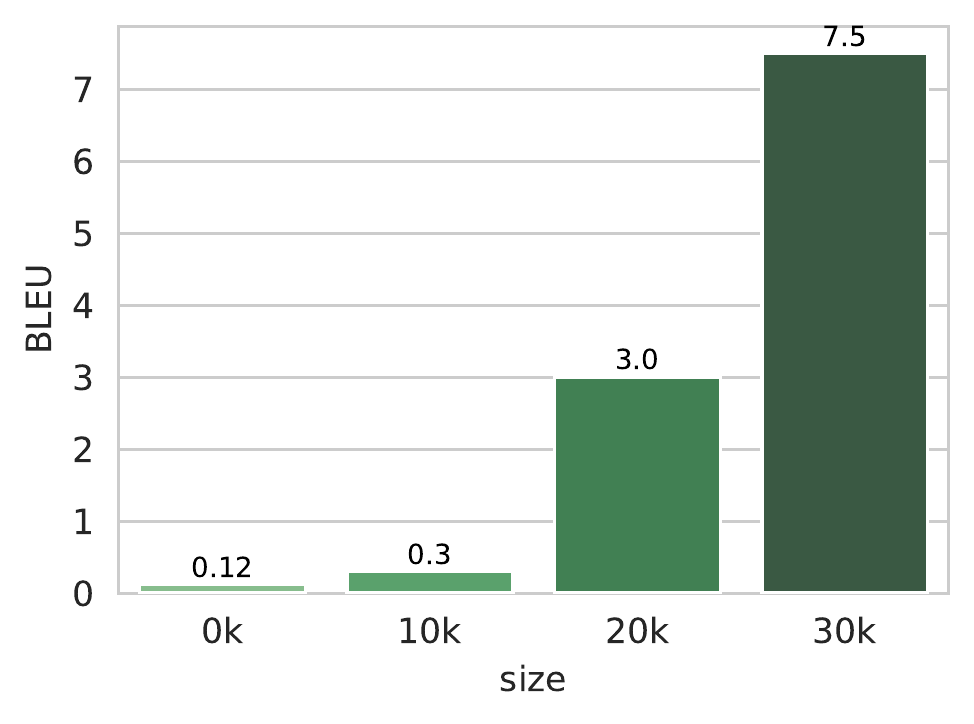}}
    \caption{English to Nepali Machine Translation results from fine-tuning on OCR-ed monolingual data collected from Nepali books corpus.
    }
    \label{fig:nepali_books}
\end{figure}

We fine-tune the model on 10k, 20k and 30k sentences and obtain significant gains in performance. The results can be seen in \Cref{fig:nepali_books}. Observe how the performance significantly increases (+7 BLEU) with the additional 30K pairs of OCR+BT data.

\subsection{The impact of OCR errors on MT}
As seen in \Cref{fig:nepali_books}, the performance of the SOTA MT model increased significantly when fine-tuned on OCR-ed data. Therefore, we want to explore in more depth what is the level of quality needed for the OCR-ed data to be useful for Machine Translation. Specifically, we want to measure the impact of OCR errors on MT performance: i) which error types affect it the most; ii) if there is an error threshold after which the OCR-ed data is detrimental to the MT model/hurts the performance; iii) if this threshold depends on data size or language.

To measure these, we first learn automatically the most frequent recognition errors that happen in each language. Then inject these errors to clean monolingual data to simulate an imperfect OCR process. Finally, we run several backtranslation experiments using the error-injected data and vary the data size and rate of OCR errors applied to the data.

\paragraph{Monolingual Data.}
We select three languages, with diverse scripts, based on their high error rates on the OCR-ed UDHR data: Khmer, Pashto and Tamil.
We apply the OCR errors on large scale monolingual data from WikiMatrix \cite{Schwenk2021WikiMatrixM1} and CC100 \cite{Wenzek2020CCNetEH, Conneau2020UnsupervisedCR}. 
To determine how the size of the monolingual data influence translation performance, we vary the data size to be 10,000 and 20,000 sentences.

\paragraph{OCR errors.}
We insert the 10 most frequent OCR errors from the best performing model on the UDHR test set.
The errors are insertions, deletions and substitutions\footnote{One interesting fact is that the most common insertion and deletion across languages is the white-space.}.
Some examples of most common character deletions and substitutions are shown in \Cref{tab:OCR_errors}.
The errors are applied randomly to the monolingual data, based on the frequency they appear in the UDHR data.
We vary the rate at which we apply the errors on the monolingual data from 0 to 20. We then measure CER. A CER of 20 means that around 20\% of the characters are incorrect.

\begin{table}[hbt!]
    \centering
    \includegraphics{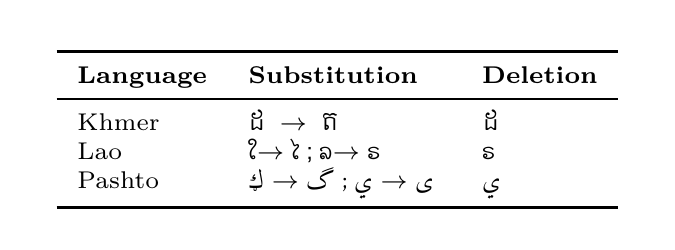}
    
    \caption{Examples of most common substitutions and deletions from UDHR OCR-ed data in Khmer, Lao and Pashto.}
    \label{tab:OCR_errors}
\end{table}

\paragraph{Backtranslation.}
We use the same MT model that we used in the Nepali experiment, the pre-trained M2M-124 model with 615M parameters from \citealt{Goyal2021TheFE}.
The source language is English and target languages are Khmer, Pashto and Tamil. 

We train a separate model for each target language. In order to measure how the OCR errors affect backtranslation performance, we run the experiments on both the initial/non OCR-ed monolingual data and the OCR-ed monolingual data.
We use the M2M-124 pre-trained model in backtranslation as following.
First, we translate the monolingual corpus into English, using the M2M-124 pre-trained model. Then, we fine-tune the model on the generated noisy English corpus and target monolingual data.
For testing the fine-tuned model we use the the Flores devtest set and for validation, the Flores dev set \cite{Goyal2021TheFE}.

\subsection{Evaluation}
\begin{figure}[hbt!]
    \centering
        \includegraphics[width=1\linewidth]{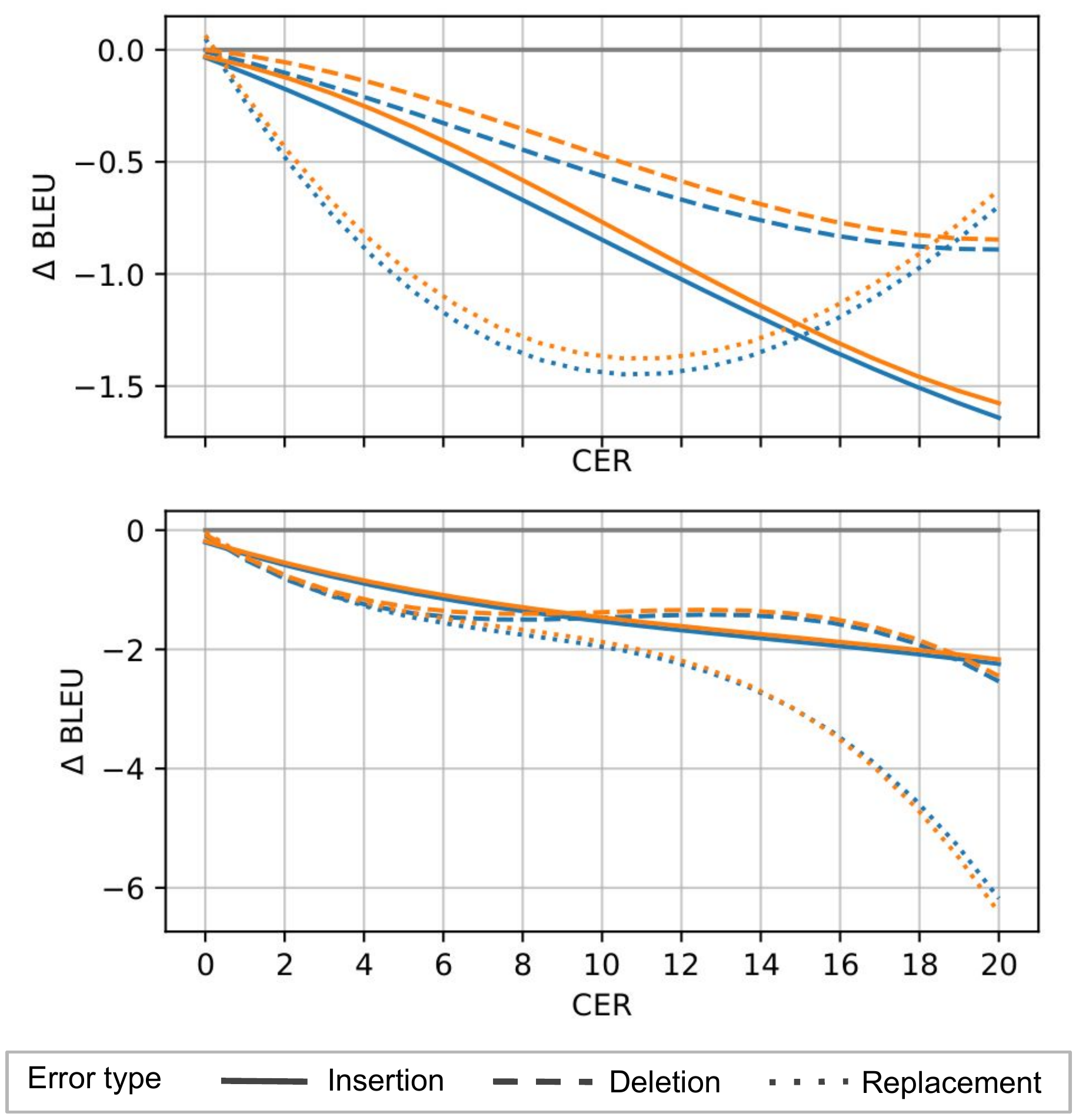}
    \caption{Ablation studies on OCR errors impact on MT performance. Upper graph (fine-tuning on 10k data) and lower graph (fine-tuning on 20k data) show the difference in BLEU scores between the M2M-124 MT model fine-tuned on OCR-ed data and the pre-trained M2M-124 MT model (shown in orange) and the difference in BLEU scores between the M2M-124 MT model fine-tuned on OCR-ed data and the M2M-124 MT model fine-tuned on non OCR-ed data (shown in blue).
    }
    \label{fig:ablations}
\end{figure}

We compare the performance of the M2M-124 fine-tuned on OCR-ed monolingual data with the M2M-124 pre-trained model and with the M2M-124 fine-tuned on initial/non OCR-ed monolingual data.
The evaluation metric used is BLEU score over tokenized text with an spm model \cite{Goyal2021TheFE}. 
The results can be seen in \Cref{fig:ablations}.

Our observations and takeaways from this ablation are the following. 

\begin{itemize}
  \setlength\itemsep{-0.1em}
     \item \textbf{Translation quality is robust to small amounts of noise}. When comparing performance of fine-tuning MT models on the OCR-ed data vs. initial/non OCR-ed, the MT performance varies per language, but on average, until CER 4\%, there is very few difference in BLEU score. Therefore, OCR-ed data with average OCR acuracy ($\leq$ 4\% CER) can be effectively used for fine-tuning MT models. Beyond that threshold, more degradation can be expected. However, in absense of any other data, noisy OCR-ed data still provides an advantage.

    \item \textbf{Replacements are more damaging than other errors.} The different types of OCR errors (insertion, deletion and replacement) have different effects on the overall MT performance. 
    On average, the replacement OCR error affects MT performance more than insertions and deletions: e.g., for fine-tuning data size 20k, until CER $\sim$10, the drop in performance caused by deletions or insertions is negligible and reaches -2 BLEU by CER 20, while replacements reduce the BLEU score much faster than the other error types ($\sim$-2 BLEU at CER 10 and -6 BLEU at CER 20). 
    Therefore, OCR-ed data with average OCR accuracy (CER $\leq$ 10) with mostly insertion and deletion errors can be effectively used for fine-tuning MT models.
    
    \item \textbf{More data results on higher or more rapid decreases in BLEU scores.} This trend is observed mostly for replacement errors. The insertions and deletions affect the OCR performance about the same amount (~-2 BLEU at CER 20) in both 10k and 20k fine-tuning data size.

\end{itemize}

\section{Conclusion}

In this paper, we proposed a new benchmark with real and synthetic data, enriched with noise, for 60 low-resource languages in low resource scripts. We group the 60 languages into groups according to their scripts and location, evaluate SOTA OCR models on our benchmark and extract their most common errors. We use the SOTA OCR errors to measure their impact on Machine Translation models by comparing the MT models fine-tuned with OCR-ed data with pre-trained MT models and MT models fine-tuned with initial/non OCR-ed data.

Our most important takeaway is that OCR-ed monolingual data improves Machine Translation (MT) through backtranslation. This augmentation is robust to most types of errors, except replacements, and in general most current OCR models produce good enough recognition to be able to train MT models, with the exception of a few scripts like Perso Arabic.

Our work paves the way for future research on data augmentation for Machine Translation based on OCR documents.

The scripts to download and process the benchmark introduced in this paper are available at \url{https://github.com/facebookresearch/flores}.

\begin{table*}
  \centering
  {
  \renewcommand{\arraystretch}{1.2}%
    \scriptsize
  \begin{tabular}{l l l R{1.50cm} R{1.5cm} R{1.50cm}  R{1.5cm}  }
  \toprule
    & & & \multicolumn{2}{c}{\textbf{Flores 101}} & \multicolumn{2}{c}{\textbf{UDHR}} \\
    \cmidrule(r){4-5}\cmidrule(l){6-7}
    \textbf{Language} & \textbf{Script} & \textbf{Group} & Tesseract & \citealt{Fujii2017SequencetoLabelSI} & Tesseract & \citealt{Fujii2017SequencetoLabelSI} \\
    \midrule
Arabic & Arabic & Perso-Arabic & 9.0 & 3.9 & 9.4 & 4.8 \\
Sorani Kurdish & Arabic & Perso-Arabic & 41.6 & 29.5 & 10.2 & 1.4 \\
Armenian & Armenian & Other & 6.4 & 0.4 & 40.6 & 39.8 \\
Bengali & Bengali & Indo-Aryan & 5.3 & 4.1 & 3.7 & 1.6 \\
Belarusian & Cyrillic & Cyrillic & 0.6 & 0.4 & 0.7 & 1.2 \\
Bulgarian & Cyrillic & Cyrillic & 0.8 & 0.2 & 0.8 & 0.8 \\
Kazakh & Cyrillic & Cyrillic & 1.2 & 0.2 & 1.3 & 1.3 \\
Kyrgyz & Cyrillic & Cyrillic & 0.8 & 0.2 & 1.9 & 3.0 \\
Macedonian & Cyrillic & Cyrillic & 0.6 & 0.2 & 0.6 & 1.5 \\
Mongolian & Cyrillic & Cyrillic & 0.2 & 0.1 & 1.8 & 1.6 \\
Russian & Cyrillic & Cyrillic & 1.0 & 0.3 & 0.5 & 1.3 \\
Serbian & Cyrillic & Cyrillic & 0.4 & 0.2 & 1.3 & 1.7 \\
Tajik & Cyrillic & Cyrillic & 1.0 & 0.2 & 2.1 & 2.9 \\
Ukrainian & Cyrillic & Cyrillic & 0.7 & 0.3 & 3.2 & 3.4 \\
Hindi & Devanagari & Indo-Aryan & 0.9 & 0.5 & 1.8 & 0.3 \\
Marathi & Devanagari & Indo-Aryan & 0.7 & 0.3 & 1.2 & 1.5 \\
Nepali & Devanagari & Indo-Aryan & 1.4 & 0.9 & 30.6 & 26.0 \\
Amharic & Ge’ez & Other & 25.3 & 3.8 & 15.1 & 45.2 \\
Georgian & Georgian & Other & 1.1 & 0.1 & 19.4 & 17.6 \\
Greek & Greek & Other & 3.0 & 0.1 & 2.5 & 0.7 \\
Gujarati & Gujarati & Indo-Aryan & 1.4 & 0.9 & 10.2 & 5.2 \\
Punjabi & Gurmukhi & Indo-Aryan & 5.0 & 2.4 & 3.1 & 2.1 \\
Japanese & Han, Hiragana, Katakana & CJK & 2.0 & 0.1 & 6.4 & 4.8 \\
Korean & Hangul & CJK & 59.8 & 1.7 & 5.4 & 3.8 \\
Chinese Simpl & Hant & CJK & 6.3 & 10.4 & 9.0 & 5.3 \\
Hebrew & Hebrew & Other & 5.2 & 4.9 & 1.3 & 1.4 \\
Khmer & Khmer & SEA & 26.1 & 9.0 & 15.9 & 12.8 \\
Lao & Lao & SEA & 17.1 & 2.6 & 67.9 & 32.4 \\
Asturian & Latin & Latin & 2.3 & 0.4 & 2.9 & 0.9 \\
Cebuano & Latin & Latin & 0.3 & 0.1 & 1.1 & 0.7 \\
Fula & Latin & Latin & 2.5 & 1.9 & 5.5 & 5.2 \\
Ganda & Latin & Latin & 0.9 & 0.1 & 1.6 & 1.1 \\
Icelandic & Latin & Latin & 0.1 & 0.1 & 28.8 & 28.6 \\
Lingala & Latin & Latin & 0.3 & 0.1 & 1.2 & 0.9 \\
Maori & Latin & Latin & 0.3 & 0.3 & 57.7 & 57.6 \\
Nyanja & Latin & Latin & 0.8 & 0.1 & 2.3 & 0.8 \\
Oromo & Latin & Latin & 3.9 & 0.2 & 2.7 & 0.7 \\
Polish & Latin & Latin & 0.1 & 0.1 & 0.6 & 0.7 \\
Portuguese (Por.) & Latin & Latin & 0.1 & 0.1 & 3.3 & 1.6 \\
Romanian & Latin & Latin & 1.4 & 0.4 & 2.0 & 1.8 \\
Shona & Latin & Latin & 0.9 & 0.1 & 1.1 & 0.8 \\
Slovak & Latin & Latin & 0.3 & 0.1 & 16.0 & 16.1 \\
Slovenian & Latin & Latin & 0.4 & 0.1 & 25.6 & 25.6 \\
Somali & Latin & Latin & 1.3 & 0.1 & 4.0 & 0.7 \\
Swahili & Latin & Latin & 0.3 & 0.1 & 0.5 & 0.7 \\
Swedish & Latin & Latin & 0.1 & 0.1 & 25.1 & 25.1 \\
Turkish & Latin & Latin & 0.2 & 0.1 & 0.6 & 0.8 \\
Umbundu & Latin & Latin & 2.8 & 1.0 & 2.5 & 1.7 \\
Uzbek & Latin & Latin & 0.1 & 0.1 & 5.2 & 5.3 \\
Vietnamese & Latin & Latin & 0.8 & 0.2 & 0.2 & 0.1 \\
Wolof & Latin & Latin & 3.6 & 0.4 & 6.1 & 2.1 \\
Zulu & Latin & Latin & 1.4 & 0.2 & 1.2 & 0.7 \\
Malayalam & Malayalam & Dradivian & 6.8 & 0.6 & 18.5 & 19.2 \\
Burmese & Myanmar & SEA & 64.6 & 9.8 & 78.3 & 1.0 \\
Pashto & Perso-Arabic & Perso-Arabic & 15.2 & 15.9 & 30.4 & 27.5 \\
Urdu & Perso-Arabic & Perso-Arabic & 4.2 & 5.6 & 53.7 & 18.9 \\
Tamil & Tamil & Dradivian & 0.9 & 0.2 & 14.1 & 11.2 \\
Kannada & Telugu-Kannada & Dradivian & 4.5 & 0.9 & 3.2 & 4.1 \\
Telugu & Telugu-Kannada & Dradivian & 3.7 & 0.7 & 32.3 & 13.9 \\
Thai & Thai & SEA & 5.0 & 1.2 & 26.9 & 9.4 \\
    \midrule
    \multicolumn{3}{c}{\textbf{Average error}} & 5.9  & 2.0  & 12.1  & 8.5 \\
 \bottomrule
  \end{tabular}
  \caption{Evaluation of SOTA models on our benchmark: CER on artificial PDFs (Flores 101) and real PDFs (UDHR), sorted alphabetically, by script and language name.}
   \label{tab:eval_SOTA}
  }
 
\end{table*}

\bibliographystyle{acl_natbib}
\bibliography{main}

\begin{thebibliography}{29}
\expandafter\ifx\csname natexlab\endcsname\relax\def\natexlab#1{#1}\fi

\bibitem[{Berg-Kirkpatrick et~al.(2013)Berg-Kirkpatrick, Durrett, and
  Klein}]{BergKirkpatrick2013UnsupervisedTO}
Taylor Berg-Kirkpatrick, Greg Durrett, and Dan Klein. 2013.
\newblock Unsupervised transcription of historical documents.
\newblock In \emph{ACL}.

\bibitem[{Bustamante et~al.(2020)Bustamante, Oncevay, and
  Zariquiey}]{Bustamante2020NoDT}
Gina Bustamante, Arturo Oncevay, and R.~Zariquiey. 2020.
\newblock No data to crawl? monolingual corpus creation from pdf files of truly
  low-resource languages in peru.
\newblock In \emph{LREC}.

\bibitem[{Chen et~al.(2021)Chen, Ma, Chen, Dong, Zhang, Pan, Wang, and
  Wei}]{chen-etal-2021-zero}
Guanhua Chen, Shuming Ma, Yun Chen, Li~Dong, Dongdong Zhang, Jia Pan, Wenping
  Wang, and Furu Wei. 2021.
\newblock \href {https://aclanthology.org/2021.emnlp-main.2} {Zero-shot
  cross-lingual transfer of neural machine translation with multilingual
  pretrained encoders}.
\newblock In \emph{Proceedings of the 2021 Conference on Empirical Methods in
  Natural Language Processing}, pages 15--26, Online and Punta Cana, Dominican
  Republic. Association for Computational Linguistics.

\bibitem[{Chiron et~al.(2017)Chiron, Doucet, Coustaty, and
  Moreux}]{Chiron2017ICDAR2017CO}
Guillaume Chiron, Antoine Doucet, Micka{\"e}l Coustaty, and Jean-Philippe
  Moreux. 2017.
\newblock Icdar2017 competition on post-ocr text correction.
\newblock \emph{2017 14th IAPR International Conference on Document Analysis
  and Recognition (ICDAR)}, 01:1423--1428.

\bibitem[{Conneau et~al.(2020)Conneau, Khandelwal, Goyal, Chaudhary, Wenzek,
  Guzm{\'a}n, Grave, Ott, Zettlemoyer, and
  Stoyanov}]{Conneau2020UnsupervisedCR}
Alexis Conneau, Kartikay Khandelwal, Naman Goyal, Vishrav Chaudhary, Guillaume
  Wenzek, Francisco Guzm{\'a}n, Edouard Grave, Myle Ott, Luke Zettlemoyer, and
  Veselin Stoyanov. 2020.
\newblock Unsupervised cross-lingual representation learning at scale.
\newblock In \emph{ACL}.

\bibitem[{Cousineau and Chartier(2010)}]{Cousineau2010OutliersDA}
Denis Cousineau and Sylvain Chartier. 2010.
\newblock Outliers detection and treatment: a review.
\newblock \emph{International journal of psychological research}, 3:58--67.

\bibitem[{Edunov et~al.(2018)Edunov, Ott, Auli, and
  Grangier}]{edunov-etal-2018-understanding}
Sergey Edunov, Myle Ott, Michael Auli, and David Grangier. 2018.
\newblock \href {https://doi.org/10.18653/v1/D18-1045} {Understanding
  back-translation at scale}.
\newblock In \emph{Proceedings of the 2018 Conference on Empirical Methods in
  Natural Language Processing}, pages 489--500, Brussels, Belgium. Association
  for Computational Linguistics.

\bibitem[{Fan et~al.(2020)Fan, Bhosale, Schwenk, Ma, El-Kishky, Goyal, Baines,
  Çelebi, Wenzek, Chaudhary, Goyal, Birch, Liptchinsky, Edunov, Grave, Auli,
  and Joulin}]{fan2020BeyondEM}
Angela Fan, Shruti Bhosale, Holger Schwenk, Zhiyi Ma, Ahmed El-Kishky,
  Siddharth Goyal, Mandeep Baines, Onur Çelebi, Guillaume Wenzek, Vishrav
  Chaudhary, Naman Goyal, Tom Birch, Vitaliy Liptchinsky, Sergey Edunov,
  Edouard Grave, Michael Auli, and Armand Joulin. 2020.
\newblock Beyond english-centric multilingual machine translation.
\newblock \emph{ArXiv}, abs/2010.11125.

\bibitem[{Fujii et~al.(2017)Fujii, Driesen, Baccash, Hurst, and
  Popat}]{Fujii2017SequencetoLabelSI}
Yasuhisa Fujii, K.~Driesen, J.~Baccash, Ash Hurst, and Ashok Popat. 2017.
\newblock Sequence-to-label script identification for multilingual ocr.
\newblock \emph{2017 14th IAPR International Conference on Document Analysis
  and Recognition (ICDAR)}, 01:161--168.

\bibitem[{Gao et~al.(2019)Gao, Zhu, Wu, Xia, Qin, Cheng, Zhou, and
  Liu}]{gao-etal-2019-soft}
Fei Gao, Jinhua Zhu, Lijun Wu, Yingce Xia, Tao Qin, Xueqi Cheng, Wengang Zhou,
  and Tie-Yan Liu. 2019.
\newblock \href {https://doi.org/10.18653/v1/P19-1555} {Soft contextual data
  augmentation for neural machine translation}.
\newblock In \emph{Proceedings of the 57th Annual Meeting of the Association
  for Computational Linguistics}, pages 5539--5544, Florence, Italy.
  Association for Computational Linguistics.

\bibitem[{Goyal et~al.(2021)Goyal, Gao, Chaudhary, Chen, Wenzek, Ju, Krishnan,
  Ranzato, Guzm{\'a}n, and Fan}]{Goyal2021TheFE}
Naman Goyal, Cynthia Gao, Vishrav Chaudhary, Peng-Jen Chen, Guillaume Wenzek,
  Da~Ju, Sanjan Krishnan, Marc'Aurelio Ranzato, Francisco Guzm{\'a}n, and
  Angela Fan. 2021.
\newblock The flores-101 evaluation benchmark for low-resource and multilingual
  machine translation.
\newblock \emph{ArXiv}, abs/2106.03193.

\bibitem[{Gupte et~al.(2021)Gupte, Romanov, Mantravadi, Banda, Liu, Khan,
  Meenal, Han, and Srinivasan}]{Gupte2021LightsCA}
Amit Gupte, Alexey Romanov, Sahitya Mantravadi, Dalitso Banda, Jianjie Liu,
  Raza Khan, Lakshmanan~Ramu Meenal, Benjamin Han, and Soundar Srinivasan.
  2021.
\newblock Lights, camera, action! a framework to improve nlp accuracy over ocr
  documents.
\newblock \emph{ArXiv}, abs/2108.02899.

\bibitem[{Hochreiter and Schmidhuber(1997)}]{hochreiter1997long}
Sepp Hochreiter and J{\"u}rgen Schmidhuber. 1997.
\newblock Long short-term memory.
\newblock \emph{Neural computation}, 9(8):1735--1780.

\bibitem[{Holley(2009)}]{Holley2009HowGC}
Rose Holley. 2009.
\newblock How good can it get? analysing and improving ocr accuracy in large
  scale historic newspaper digitisation programs.
\newblock \emph{D Lib Mag.}, 15.

\bibitem[{Ingle et~al.(2019)Ingle, Fujii, Deselaers, Baccash, and
  Popat}]{Ingle2019ASH}
R.~Reeve Ingle, Yasuhisa Fujii, Thomas Deselaers, Jonathan Baccash, and Ashok
  Popat. 2019.
\newblock A scalable handwritten text recognition system.
\newblock \emph{2019 International Conference on Document Analysis and
  Recognition (ICDAR)}, pages 17--24.

\bibitem[{Liu et~al.(2021)Liu, Kusner, and
  Blunsom}]{liu-etal-2021-counterfactual}
Qi~Liu, Matt Kusner, and Phil Blunsom. 2021.
\newblock \href {https://doi.org/10.18653/v1/2021.naacl-main.18}
  {Counterfactual data augmentation for neural machine translation}.
\newblock In \emph{Proceedings of the 2021 Conference of the North American
  Chapter of the Association for Computational Linguistics: Human Language
  Technologies}, pages 187--197, Online. Association for Computational
  Linguistics.

\bibitem[{Pradhan et~al.(2012)Pradhan, Moschitti, Xue, Uryupina, and
  Zhang}]{Pradhan2012CoNLL2012ST}
Sameer Pradhan, Alessandro Moschitti, Nianwen Xue, O.~Uryupina, and Yuchen
  Zhang. 2012.
\newblock Conll-2012 shared task: Modeling multilingual unrestricted
  coreference in ontonotes.
\newblock In \emph{EMNLP-CoNLL Shared Task}.

\bibitem[{Rigaud et~al.(2019)Rigaud, Doucet, Coustaty, and
  Moreux}]{Rigaud2019ICDAR2C}
Christophe Rigaud, Antoine Doucet, Micka{\"e}l Coustaty, and Jean-Philippe
  Moreux. 2019.
\newblock Icdar 2019 competition on post-ocr text correction.
\newblock \emph{2019 International Conference on Document Analysis and
  Recognition (ICDAR)}, pages 1588--1593.

\bibitem[{Rijhwani et~al.(2020{\natexlab{a}})Rijhwani, Anastasopoulos, and
  Neubig}]{Rijhwani2020OCRPF}
Shruti Rijhwani, Antonios Anastasopoulos, and Graham Neubig.
  2020{\natexlab{a}}.
\newblock Ocr post-correction for endangered language texts.
\newblock \emph{ArXiv}, abs/2011.05402.

\bibitem[{Rijhwani et~al.(2020{\natexlab{b}})Rijhwani, Anastasopoulos, and
  Neubig}]{rijhwani-etal-2020-ocr}
Shruti Rijhwani, Antonios Anastasopoulos, and Graham Neubig.
  2020{\natexlab{b}}.
\newblock \href {https://doi.org/10.18653/v1/2020.emnlp-main.478} {{OCR} {P}ost
  {C}orrection for {E}ndangered {L}anguage {T}exts}.
\newblock In \emph{Proceedings of the 2020 Conference on Empirical Methods in
  Natural Language Processing (EMNLP)}, pages 5931--5942, Online. Association
  for Computational Linguistics.

\bibitem[{Sang and Meulder(2003)}]{Sang2003IntroductionTT}
E.~T.~K. Sang and F.~D. Meulder. 2003.
\newblock Introduction to the conll-2003 shared task: Language-independent
  named entity recognition.
\newblock In \emph{CoNLL}.

\bibitem[{Schulz and Kuhn(2017)}]{Schulz2017MultimodularDO}
Sarah Schulz and Jonas Kuhn. 2017.
\newblock Multi-modular domain-tailored ocr post-correction.
\newblock In \emph{EMNLP}.

\bibitem[{Schwenk et~al.(2021)Schwenk, Chaudhary, Sun, Gong, and
  Guzm{\'a}n}]{Schwenk2021WikiMatrixM1}
Holger Schwenk, Vishrav Chaudhary, Shuo Sun, Hongyu Gong, and Francisco
  Guzm{\'a}n. 2021.
\newblock Wikimatrix: Mining 135m parallel sentences in 1620 language pairs
  from wikipedia.
\newblock \emph{ArXiv}, abs/1907.05791.

\bibitem[{Sennrich et~al.(2016)Sennrich, Haddow, and
  Birch}]{Sennrich2016ImprovingNM}
Rico Sennrich, B.~Haddow, and Alexandra Birch. 2016.
\newblock Improving neural machine translation models with monolingual data.
\newblock \emph{ArXiv}, abs/1511.06709.

\bibitem[{Smith(2007{\natexlab{a}})}]{tesseract}
R.~Smith. 2007{\natexlab{a}}.
\newblock \href {https://doi.org/10.1109/ICDAR.2007.4376991} {An overview of
  the tesseract ocr engine}.
\newblock In \emph{Ninth International Conference on Document Analysis and
  Recognition (ICDAR 2007)}, volume~2, pages 629--633.

\bibitem[{Smith(2007{\natexlab{b}})}]{Smith2007AnOO}
R.~Smith. 2007{\natexlab{b}}.
\newblock An overview of the tesseract ocr engine.
\newblock \emph{Ninth International Conference on Document Analysis and
  Recognition (ICDAR 2007)}, 2:629--633.

\bibitem[{Wenzek et~al.(2020)Wenzek, Lachaux, Conneau, Chaudhary, Guzm'an,
  Joulin, and Grave}]{Wenzek2020CCNetEH}
Guillaume Wenzek, Marie-Anne Lachaux, Alexis Conneau, Vishrav Chaudhary,
  Francisco Guzm'an, Armand Joulin, and Edouard Grave. 2020.
\newblock Ccnet: Extracting high quality monolingual datasets from web crawl
  data.
\newblock \emph{ArXiv}, abs/1911.00359.

\bibitem[{Wick et~al.(2020)Wick, Reul, and Puppe}]{wick_calamari_2020}
Christoph Wick, Christian Reul, and Frank Puppe. 2020.
\newblock Calamari - {A} {High}-{Performance} {Tensorflow}-based {Deep}
  {Learning} {Package} for {Optical} {Character} {Recognition}.
\newblock \emph{Digital Humanities Quarterly}, 14(1).

\bibitem[{Zhang et~al.(2020)Zhang, Williams, Titov, and
  Sennrich}]{zhang-etal-2020-improving}
Biao Zhang, Philip Williams, Ivan Titov, and Rico Sennrich. 2020.
\newblock \href {https://doi.org/10.18653/v1/2020.acl-main.148} {Improving
  massively multilingual neural machine translation and zero-shot translation}.
\newblock In \emph{Proceedings of the 58th Annual Meeting of the Association
  for Computational Linguistics}, pages 1628--1639, Online. Association for
  Computational Linguistics.

\end{thebibliography}

\end{document}